%% file: acl_latex.tex
\documentclass[11pt]{article}

\usepackage[final]{acl}

\usepackage{etoolbox}
\makeatletter
\patchcmd{\@outputdblcol}
  {\normalsize\baselineskip\z@skip\lineskip\z@skip}
  {\normalsize\baselineskip\z@skip\lineskip\z@skip}
  {}{}
\makeatother

\usepackage{times}
\usepackage{latexsym}
\usepackage{tcolorbox}
\usepackage{amsmath}
\usepackage[T1]{fontenc}

\usepackage[utf8]{inputenc}

\usepackage{microtype}
\usepackage{enumitem}
\usepackage{inconsolata}

\usepackage{graphicx}

\usepackage{booktabs}
\usepackage{multirow}
\usepackage{xcolor}

\newcommand{\datasetname}{\textsc{DivanBench}}

\title{Unmasking the Factual-Conceptual Gap in Persian Language Models}

\author{
  Alireza Sakhaeirad\thanks{Correspondence to: \texttt{alireza.sakhaeirad@epfl.ch} \\} \\
  EPFL \\
  \And
  Ali Ma'manpoosh \\
  University of Isfahan \\
  \And
  Arshia Hemmat \\
  University of Oxford \\
}
\begin{document}
\maketitle

\begin{abstract}
While emerging Persian NLP benchmarks have expanded into pragmatics and politeness, they rarely distinguish between memorized cultural facts and the ability to reason about implicit social norms. We introduce \datasetname, a diagnostic benchmark focused on superstitions and customs, arbitrary, context-dependent rules that resist simple logical deduction. Through 315 questions across three task types (factual retrieval, paired scenario verification, and situational reasoning), we evaluate seven Persian LLMs and reveal three critical failures: {most models exhibit severe acquiescence bias, correctly identifying appropriate behaviors but failing to reject clear violations}; {continuous Persian pretraining amplifies this bias rather than improving reasoning, often degrading the model's ability to discern contradictions}; and {all models show a 21\% performance gap between retrieving factual knowledge and applying it in scenarios}. These findings demonstrate that cultural competence requires more than scaling monolingual data, as current models learn to mimic cultural patterns without internalizing the underlying schemas.\footnote{Dataset publicly available \href{https://huggingface.co/datasets/divanbench/divanbench}{huggingface.co/datasets/divanbench/divanbench}}

\end{abstract}

\input{Sections/introduction}

\input{Sections/background}
\input{Sections/dataset}
\input{Sections/experiments}

\section{Conclusion}
\input{Sections/conclusion}

\newpage
\input{Sections/limitation}

\newpage
\bibliography{custom}

\newpage
\appendix

\input{Appendix/dataset_details}
\input{Appendix/experiment_details}

\end{document}

%% file: Sections/introduction.tex
\section{Introduction}
\label{sec:introduction}

\input{Tables/cultral_examples}

If you offer a Persian guest food, they will refuse. If you offer again, they will refuse. Only on the third offer, the ``real'' one, will they accept. This three-iteration ritual, called \textit{taarof}, is immediately obvious to any Iranian child raised in the culture. But can a language model, trained on billions of Persian tokens, distinguish genuine \textit{taarof} from a cultural violation? Our results suggest: not really.

\begin{figure}
    \centering
    \includegraphics[width=.95\linewidth]{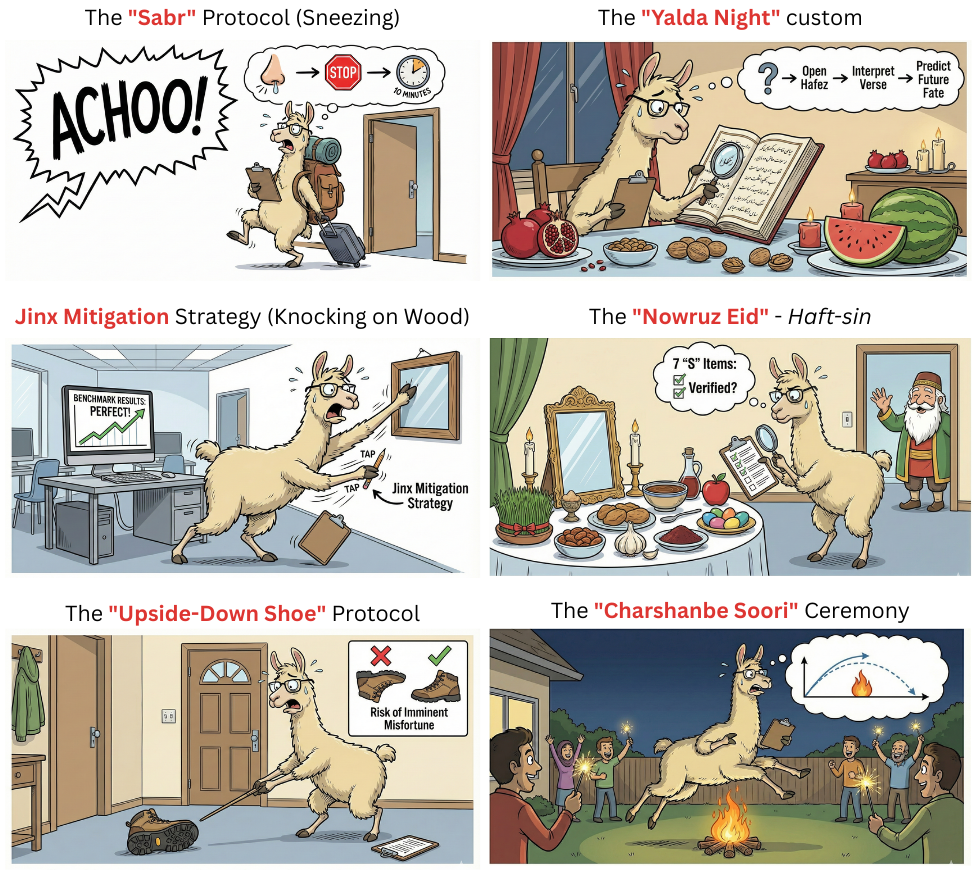}
    \caption{Sample Persian cultural concepts from the benchmark, spanning superstitions, traditions, and taboos.}
    \label{fig:demo}
\end{figure}

The rapid advancement of large language models (LLMs) has sparked significant interest in their multilingual capabilities, including Persian language processing. Current Persian LLM benchmarks primarily focus on factual knowledge retrieval, translation quality, and linguistic tasks such as sentiment analysis and named entity recognition~\cite{khashabi2020parsinlu}. However, these evaluations critically overlook a dimension central to Persian communication: the ability to reason about {implicit cultural concepts}, particularly superstitions and customs, that govern appropriate behavior in context-dependent social situations.

Persian culture embeds meaning across multiple historical layers: ancient Zoroastrian beliefs (\textit{Chaharshanbe Suri} fire-jumping, \textit{esfand} burning for evil eye protection), Islamic traditions (\textit{nazri} vow offerings, \textit{ghorbani} sacrifice), complex social etiquette systems (\textit{taarof}, \textit{jang-e hesab}), rich folk cosmology featuring supernatural beings (\textit{jinn}, \textit{div}, \textit{pari}, \textit{bakhtak}), widespread superstitious practices (whistling at night attracts evil), and elaborate life-cycle ceremonies with highly specific ritualistic requirements. These concepts are not facts to be retrieved from a knowledge base; they are {cultural schemas}~\cite{shore1996culture,strauss1992models} requiring understanding of context-dependent social dynamics, implicit power relations, and culturally-specific rules of appropriateness that are obvious to cultural insiders but opaque to pattern-matchers.

\subsection{Motivation: Why Superstitions and Customs?}

Superstitions and customs provide particularly demanding testbeds for cultural reasoning because they often lack logical justification-, unlike social etiquette, which has pragmatic benefits, superstitions require acceptance of culturally-transmitted beliefs without empirical grounding. Furthermore, they exhibit context-dependence: the same action (such as sweeping) is neutral by day but taboo at night. These practices involve implicit rules that are typically corrected through social feedback rather than explicit instruction, and Persian superstitions notably reflect historical syncretism, layering Zoroastrian, Islamic, and folk beliefs in complex ways. Recent work on Korean superstitions~\cite{kim-lee-2025-nunchi} demonstrated that culture-bound phenomena reveal limitations invisible in standard benchmarks. We extend this insight to Persian, where superstitions permeate daily life and provide rich signal for distinguishing genuine cultural understanding from keyword-matching.

\subsection{Research Questions, Contributions, and Key Findings}
\label{subsec:rq_contrib_findings}

This work (illustrated in Fig.~\ref{fig:demo}) investigates whether Persian-capable Large Language Models (LLMs) internalize \textit{cultural schemas}, the implicit social logic mapping context to action, or merely reproduce surface-level \textit{cultural facts} and stereotypes. The investigation focuses on three core dimensions: the propensity for models to exhibit acquiescence bias through pattern-matching, the impact of continuous Persian pretraining on reasoning versus fluency, and the capacity for factual knowledge to be operationalized into scenario-based application. By distinguishing between fluent cultural expression and robust cultural reasoning, this work provides a framework for evaluating cultural competence in low-resource linguistic environments.

Central to this evaluation is the introduction of \datasetname, a diagnostic framework covering 81 concepts across superstitions, customs, and social etiquette, instantiated through 315 questions. The benchmark adopts a three-level architecture designed to isolate distinct layers of cultural competence: {Factual MCQ} for baseline knowledge retrieval, {Binary Belief Verification} for testing discernment through paired positive/negative scenarios, and {Scenario-Based MCQ} for evaluating multi-step inference in complex social contexts. By utilizing a paired design, contrasting behavior that aligns with Persian customs against plausible but culturally inappropriate alternatives, this methodology effectively quantifies acquiescence bias. Furthermore, the study isolates the effects of monolingual adaptation by comparing a base model, Llama 3.1-8B, its Persian-adapted variant  Dorna2-8B, revealing how specialized pretraining affects the transition from surface fluency to logical consistency.

Our evaluation of 7 state-of-the-art models yields the following key findings:

\begin{itemize}
    \item \textbf{The acquiescence Trap:} 6 of 7 models exhibit significant asymmetry in reasoning; while they identify appropriate behaviors with 84\%--92\% accuracy, they fail to reject cultural violations at rates between 19\% and 48\%, suggesting a reliance on pattern-matching over logic.
    \item \textbf{The Persian Pretraining Paradox:} Continuous pretraining can inadvertently degrade critical reasoning. Llama 3.1-8B’s rejection accuracy of 73\% dropped to 30\% after Persian-specific adaptation (Dorna2), representing a 43-percentage-point decline.
    \item \textbf{The Factual--Conceptual Gap:} Performance decreases by an average of 21\% when transitioning from factual retrieval to scenario-based reasoning, demonstrating that memorized cultural facts do not reliably translate into functional cultural schemas.
    \item \textbf{Distinct Learning Targets:} The results provide empirical evidence that cultural facts and schemas behave as independent objectives, suggesting that cultural competence requires training mechanisms beyond standard data scaling or monolingual adaptation.
\end{itemize}

The remainder of this paper is structured as follows: Section~\ref{sec:related} reviews related work on cultural evaluation and Persian NLP; Section~\ref{sec:dataset} describes our benchmark design, cultural concept coverage, and data collection methodology. Section~\ref{sec:experiments} details our experimental setup including model selection, evaluation protocol, and experimental design. Subsection~\ref{sec:results} presents our four main findings on acquiescence bias, pretraining effects, factual-conceptual gaps, and model scaling. Subsection~\ref{sec:discussion} analyzes the mechanisms behind these findings and their implications for multilingual NLP. Finally, Section~\ref{sec:conclusion} concludes and discusses future directions for cultural reasoning research. Appendix \ref{sec:appendixData} gives more insight into the data and presents more examples, and appendix \ref{sec:extra-exp} adds some additional detail on system prompt variations, experiments, and results.

%% file: Tables/cultral_examples.tex
\begin{table*}[t]
\centering
\small
\begin{tabular}{l r p{10cm}}
\hline
\textbf{Category} & \textbf{Example Concepts} \\
\hline
Social Etiquette & \textit{Taarof}, \textit{Jang-e Hesab}, Doorway Deference, Three-Times Rule, \textit{Pishkesh}, \textit{Shirini} \\
Nowruz Traditions & \textit{Haft Sin}, \textit{Chaharshanbe Suri}, \textit{Haji Firuz}, \textit{Samanu Pazan}, \textit{Khaneh Tekani}, \textit{Eidi} \\
Supernatural Beings & \textit{Jinn}, \textit{Div}, \textit{Pari}, \textit{Bakhtak}, \textit{Hamzad}, \textit{Al} \\
Apotropaic Rituals & \textit{Nazar} amulet, \textit{Esfand} burning, Salt circle, \textit{Bismillah} invocations \\
Wedding Ceremonies & \textit{Sofreh Aghd}, Knife Dance, \textit{Kuzeh Shekani}, Honey ritual \\
Taboos & Whistling at night, Sweeping at night, Stepping on bread, Shoe taboos, Nail cutting restrictions \\
Divination/Omens & \textit{Fal-e Hafez}, \textit{Fal-Gush}, Dream interpretation, Itchy palms, Ear ringing \\
\hline
\end{tabular}
\caption{Distribution of cultural concepts across categories. Concepts span ancient Zoroastrian beliefs, Islamic traditions, social etiquette systems, folk cosmology, life-cycle ceremonies, taboos, and divination practices.}
\label{tab:concepts}
\end{table*}

%% file: Sections/background.tex
\section{Related Work}
\label{sec:related}

\subsection{Persian LLM Evaluation}
Most Persian NLP benchmarks have traditionally emphasized linguistic competence and factual or domain knowledge. Representative resources include \textsc{ParsiNLU} for broad NLU coverage~\cite{khashabi2020parsinlu}, \textsc{FarsTail} for Persian natural language inference~\cite{amirkhani2021farstailpersiannaturallanguage}, and \textsc{PersianMedQA} for bilingual medical question answering~\cite{kalahroodi2025persianmedqaevaluatinglargelanguage}. Recent efforts broaden evaluation toward general knowledge and multimodal educational assessment, including the Khayyam Challenge (PersianMMLU)~\cite{ghahroodi2024khayyamchallengepersianmmlullm} and MEENA (PersianMMMU)~\cite{ghahroodi2025meenapersianmmmumultimodalmultilingualeducational}. In parallel, a newer line of work targets cultural and pragmatic competence more directly: \textsc{PerCul} uses story-driven scenarios to probe cultural understanding in Persian~\cite{monazzah2025perculstorydrivenculturalevaluation}, \textsc{TaarofBench} evaluates the Persian politeness system and its nuanced social dynamics~\cite{sadr2025politelyinsistllmlearn}, and \textsc{ELAB} benchmarks Persian-relevant alignment and safety dimensions~\cite{pourbahman2025elabextensivellmalignment}. Together, these works show a shift from ``language + facts'' evaluation toward culturally grounded pragmatics, though the latter remains less standardized and comparatively underexplored.

\subsection{Cultural Competence in LLMs}
Beyond Persian, substantial evidence shows that LLMs often fail to capture culture-specific, context-dependent norms and may default to dominant viewpoints or produce plausible-but-shallow cultural explanations~\cite{zhang2025culturescopedimensionallensprobing,dai2025wordworldevaluatemitigate,durmus2024measuringrepresentationsubjectiveglobal,hossain2025craftexplanationbasedframeworkevaluating}. Cross-cultural evaluations similarly report performance disparities between high-resource and low-resource cultures and uneven representation of global perspectives~\cite{cao2023culturellm,durmus2023globalopinionqa}. Related research in social/pragmatic reasoning and perspective-taking further indicates weaknesses on tasks that require implicit modeling of beliefs and intentions~\cite{sap2020socialiqa,sap2022neural}. Particularly relevant are cultural testbeds based on superstitions (e.g., \textsc{Nunchi-Bench}), where models may handle surface facts but struggle with situational reasoning~\cite{kim-lee-2025-nunchi}.

Methodologically, evaluation can be confounded by systematic response biases: LLMs exhibit position bias in multiple-choice settings and related tendencies such as acquiescence bias/``yes-saying''~\cite{zheng2023large,wang2023position,wallace2019trick,si2022prompting}. Prior work mitigates these effects via prompt design and calibration~\cite{ko2023instruction,si2022prompting}, while alternative evaluation designs aim to measure bias explicitly. Finally, cognitive theories distinguish explicit cultural facts from implicit cultural schemas learned through repeated, socially situated experience~\cite{shore1996culture,strauss1992models}. This distinction aligns with arguments that text-only training may be insufficient for grounded understanding~\cite{bisk2020experience} and echoes the classic observations that robust common sense reasoning is difficult to recover from text statistics alone~\cite{levesque2012winograd}.

\subsection{Positioning Our Work}

Our work uniquely combines: (1) comprehensive coverage of Persian superstitions and customs (more than 80 concepts across 7 domains); (2) explicit bias measurement through paired positive/negative design; (3) direct comparison isolating pretraining effects (Llama3.1 \cite{dubey2024llama} vs. Dorna2 \cite{dorna}); and (4) theoretical grounding distinguishing cultural facts from schemas. While prior work addresses individual aspects (Persian-related factual retrieval tests, cultural evaluation, low-resource pretraining), we provide integrated analysis revealing systematic limitations in current approaches to cultural competence in Persian LLMs.

%% file: Sections/dataset.tex
\section{Benchmark Design and Data Collection}
\label{sec:dataset}

We introduce \datasetname, a benchmark for evaluating cultural reasoning in Persian LLMs. Our design philosophy centers on distinguishing genuine cultural understanding from pattern-matching to plausible-sounding Persian text.

\subsection{Design Principles}

We evaluate models across three complementary tasks isolating different aspects of cultural competence:

\paragraph{Factual Multiple-Choice Questions} The first data type consists of 100 Factual MCQs designed to test whether models can accurately retrieve fundamental information regarding Persian culture, history, geography, and traditions. These questions serve as a critical baseline control, identifying whether a model possesses the raw cultural knowledge necessary for more complex tasks. Success in this category indicates a robust factual foundation, while failure suggests a lack of exposure to Persian-specific data during the model's pre-training phase.

\paragraph{Binary Belief Verification} The second task type comprises 162 statements categorized into matched pairs for 81 distinct cultural concepts. Each pair includes a "positive" scenario indicating a person following the Persian customs and a "negative" scenario where the subject acts against those customs in a way that remains plausible in a non-Persian context. This paired design is specifically intended to measure acquiescence bias and discernment; models that rely on simple pattern matching often show an asymmetric performance, erroneously accepting culturally inappropriate behaviors while correctly identifying appropriate ones.

\paragraph{Scenario-Based Multiple-Choice Questions} The final data type includes 53 Scenario-Based MCQs that challenge models to apply cultural reasoning to complex, novel social situations. These scenarios require intricate inferences about social hierarchies, contextual appropriateness, and interpersonal nuances that are not explicitly taught. By focusing on concepts typically acquired through lived experience and social interaction, this task isolates the model's ability to move beyond rote memorization and demonstrate authentic cultural competence in dynamic settings.
\begin{tcolorbox}[colback=blue!5, colframe=blue!40!black, title=Example: Scenario-Based MCQ, fonttitle=\bfseries, top=2mm, bottom=2mm, left=3mm, right=3mm]

\textit{Arman has a sore throat and a high fever. His mother brings him soup and tea but strictly warns him not to eat the fresh melon sitting on the counter. Arman argues that the fruit is full of vitamins and will help him recover. Why does the mother forbid Arman from eating the melon while he has a cold?}

\begin{itemize}[topsep=2pt, itemsep=0pt, leftmargin=15pt]
\item[(A)] Melon is \textit{Sard} (Cold); adding cold fruit while sick will freeze the lungs and prolong recovery
\item[(B)] Melon is \textit{Garm} (Hot); eating it during a fever will ignite the blood, spreading infection faster
\item[(C)] Folk belief holds melon aroma attracts night-illness spirits, preventing medicine from working
\item[(D)] Hygiene precaution: high sugar content feeds bacteria in the throat, causing permanent voice loss
\end{itemize}

\textbf{Expected}: (A) --- requires understanding the classical Persian food classification system (\textit{Sard-Garm}) and traditional Persian medicine, which lacks modern medical evidence.

\end{tcolorbox}

\paragraph{Author-Generated Content and Quality Control} 
All questions were drafted and refined through multiple rounds of author review based on cultural knowledge acquired via lived experience in Iranian society. This process ensures authenticity by reflecting real social practices rather than stereotypes, focusing on "insider knowledge" that is intuitive to cultural members but requires active reasoning from outsiders. We prioritized implicit concepts typically learned through social interaction, rather than explicit instruction. By ensuring unambiguity for insiders while maintaining plausible alternatives for outsiders, the dataset enables a diagnostic evaluation of cultural competence that distinguishes genuine understanding from simple pattern matching.

\subsection{Cultural Concept Coverage}
Our benchmark spans seven major cultural domains, summarized in Table~\ref{tab:concepts} and detailed in Appendix~\ref{app:concepts}. These domains encompass superstitions and omens such as divination practices (\textit{Fal-e Hafez}, \textit{Fal-Gush}), dream interpretation, and bodily omens like itchy palms or ear ringing, alongside apotropaic rituals like burning \textit{esfand} to ward off the evil eye, wearing \textit{nazar} amulets, and \textit{Bismillah} invocations. Additionally, the dataset covers taboos governing daily life, including restrictions on whistling or sweeping at night, and prohibitions against stepping on bread.

%% file: Sections/experiments.tex
\section{Experiments}
\label{sec:experiments}
\subsection{Setup}

\paragraph{Model Selection}
We evaluate 7 models from the Open Persian LLM Leaderboard~\cite{openpersianlb2024} with similar parameter counts to ensure fair comparison. All models are in the 7--12B parameter range, representing state-of-the-art performance for Persian language tasks: Aya-8B~\cite{cohere2024aya}, Dorna2-8B~\cite{dorna}, Gemma2-9B and Gemma3-12B~\cite{team2023gemini}, Llama3.1-8B~\cite{dubey2024llama}, and Qwen2-7B and Qwen2.5-7B~\cite{qwen2024}. The critical comparison is Llama3.1-8B (base) versus Dorna2-8B (Persian-adapted through continuous pretraining), which isolates the effect of Persian-focused pretraining while controlling for all other modeling choices. (For detailed description about models see App.~\ref{app:models})

\paragraph{Evaluation Protocol}
Following best practices for reproducible LLM evaluation~\cite{chang2024survey}, we use temperature of $0.1$ and top-p sampling set to $0.9$ with fixed random seeds. For answer extraction, we employ GPT-4o-mini as a systematic extraction agent~\cite{zheng2023large}, which parses model outputs to identify selected options (A/B/C/D) for multiple-choice questions and yes/no responses for binary questions. To account for prompt sensitivity~\cite{si2022prompting}, each question is evaluated with 5 diverse system prompts that vary in phrasing while maintaining semantic equivalence. All prompts instruct models to respond as Iranian cultural insiders to ensure fair evaluation conditions. We report mean accuracy and standard deviation across prompt variations. To see exact system prompts, see {\ref{app:prompts}}.

We compute three complementary metrics: (1) \textbf{Accuracy} measuring percentage of correct responses per task type; (2) \textbf{Acquiescence bias} quantifying the difference between acceptance rates for positive scenarios and rejection rates for negative scenarios in binary tasks, where high positive bias indicates pattern-matching rather than reasoning; and (3) \textbf{Factual-Conceptual Gap} measuring performance difference between factual retrieval and scenario-based reasoning. These metrics provide multi-dimensional assessment of cultural competence beyond standard accuracy reporting.

\paragraph{Experimental Design}

We conduct three complementary experiments to systematically evaluate cultural reasoning in Persian LLMs:

\begin{enumerate}

\item \textbf{Acquiescence bias Measurement (Experiment 1).} We compare model performance on paired positive and negative binary scenarios testing identical cultural concepts. This paired design directly measures acquiescence bias as $B = \text{Acc}_{\text{pos}} - \text{Acc}_{\text{neg}}$, where $\text{Acc}_{\text{pos}}$ is accuracy on appropriate behavior and $\text{Acc}_{\text{neg}}$ is accuracy on violations. Models with high positive bias demonstrate pattern-matching to cultural keywords rather than genuine reasoning.

\item \textbf{Pretraining Effects (Experiment 2).} We conduct a controlled comparison between Llama3.1-8B (base model $M_{\text{base}}$) and Dorna2-8B (Persian-adapted model $M_{\text{adapted}}$). This matched-pair design isolates the effect of Persian pretraining by comparing performance changes $\Delta = \text{Acc}(M_{\text{adapted}}) - \text{Acc}(M_{\text{base}})$ across all task types, revealing whether continuous pretraining improves cultural reasoning or reinforces surface-level pattern-matching.

\item \textbf{Knowledge Transfer (Experiment 3).} We analyze the factual-conceptual gap as $G = \text{Acc}_{\text{factual}} - \text{Acc}_{\text{scenario}}$, where $\text{Acc}_{\text{factual}}$ measures retrieval and $\text{Acc}_{\text{scenario}}$ measures cultarul reasoning. Large consistent gaps suggest cultural facts and schemas engage different cognitive mechanisms.
\end{enumerate}
\input{Sections/results}

\subsection{Analysis}
\input{Sections/discussion}

%% file: Sections/results.tex
\input{Tables/main_table}

\subsection{Results}
\label{sec:results}

We present results across all evaluation dimensions, revealing systematic patterns in how Persian language models handle cultural reasoning. Table~\ref{tab:main_results} shows complete performance across all tasks (See table \ref{tab:binary_performance} for detailed results of binary tests).

\paragraph{Finding 1: The acquiescence Trap}

Most models exhibit severe acquiescence bias on cultural questions, accepting plausible statements far more readily than rejecting violations. This asymmetry reveals that models are pattern-matching to culturally-themed Persian text rather than reasoning about appropriateness. Extreme cases include Qwen2-7B (84\% acceptance vs. 19\% rejection, bias = +65\%), Gemma3-12B (92\% acceptance vs. 31\% rejection, bias = +61\%), and Dorna2-8B (91\% acceptance vs. 30\% rejection, bias = +61\%). These models correctly identify culturally appropriate behavior when presented positively but fail to recognize violations of the same concepts when presented negatively. for instance, accepting both "Standing up when elders arrive" (correct code of respect) and "Remaining seated for comfort when elders arrive" (violates code of respect). The only exception is Llama3.1-8B, showing opposite bias (73\% rejection vs. 63\% acceptance, bias = -10\%), suggesting an skeptical tone in this model.

\begin{figure}[t]
\centering
\includegraphics[width=\columnwidth]{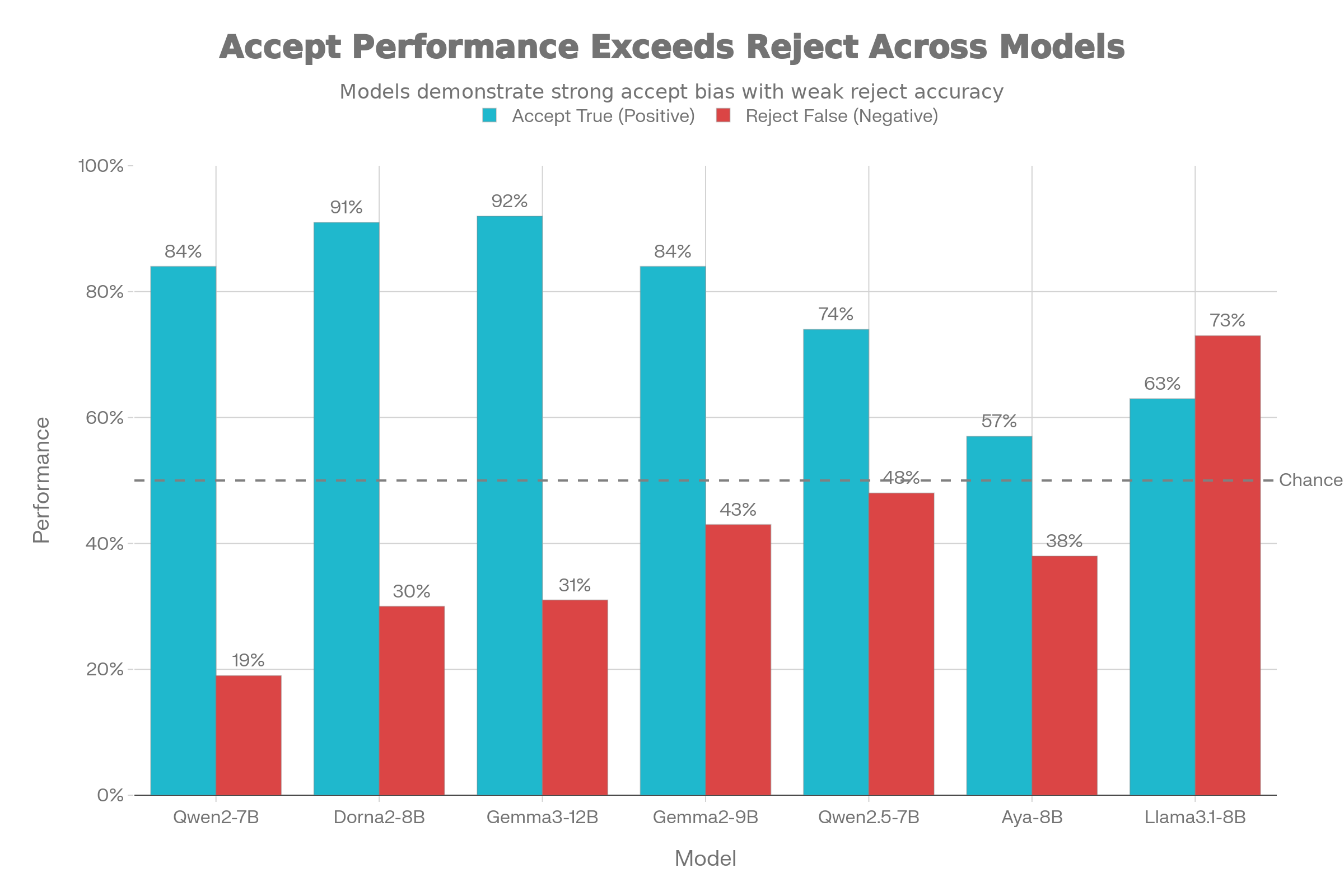}
\caption{acquiescence bias across models. Most models accept true cultural statements readily (Blue) but fail to reject false ones (Red), indicating pattern-matching rather than reasoning.}
\label{fig:agreeing_bias}
\end{figure}

\input{Tables/Llama_vs_Dorna}

\paragraph{Finding 2: The Persian Pretraining Paradox}

Comparing Llama3.1-8B (base) with Dorna2-8B (Persian-tuned) in table \ref{tab:llama_vs_dorna} reveals counterintuitive effects of continuous pretraining on cultural reasoning. Rejection accuracy collapsed dramatically from 73\% to 30\% (-43 percentage points), while acceptance accuracy soared from 63\% to 91\% (+28pp). Notably, factual knowledge slightly declined (80\% to 73\%) and scenario reasoning remained essentially flat (53\% to 54\%), indicating that continuous Persian pretraining taught the model to recognize cultural patterns without improving reasoning about cultural logic. This suggests that more Persian data reinforced surface-level pattern-matching over deep cultural understanding. Dorna2 became more ``culturally compliant'' but less discerning, accepting any plausible-sounding cultural scenario regardless of correctness. This finding challenges the common assumption in low-resource NLP that monolingual pretraining universally improves understanding.

\paragraph{Finding 3: The Factual-Conceptual Gap}

All models show consistent performance degradation from factual to scenario-based tasks, with an average gap of 21 percentage points. Models can retrieve facts about Persian culture ("What is \textit{taarof}?") but struggle to apply cultural logic to novel scenarios ("In this situation, would \textit{taarof} be appropriate?"). This gap demonstrates that knowing that a concept exists differs fundamentally from knowing when it appropriately applies. The gap persists even for the best-performing model (Gemma3-12B shows -21\%), suggesting this is a fundamental limitation rather than a simple capacity issue. This evidence supports the cognitive anthropology distinction between cultural facts (retrievable from text) and cultural schemas (requiring embodied social learning).

\begin{figure}[t]
\centering
\includegraphics[width=\columnwidth]{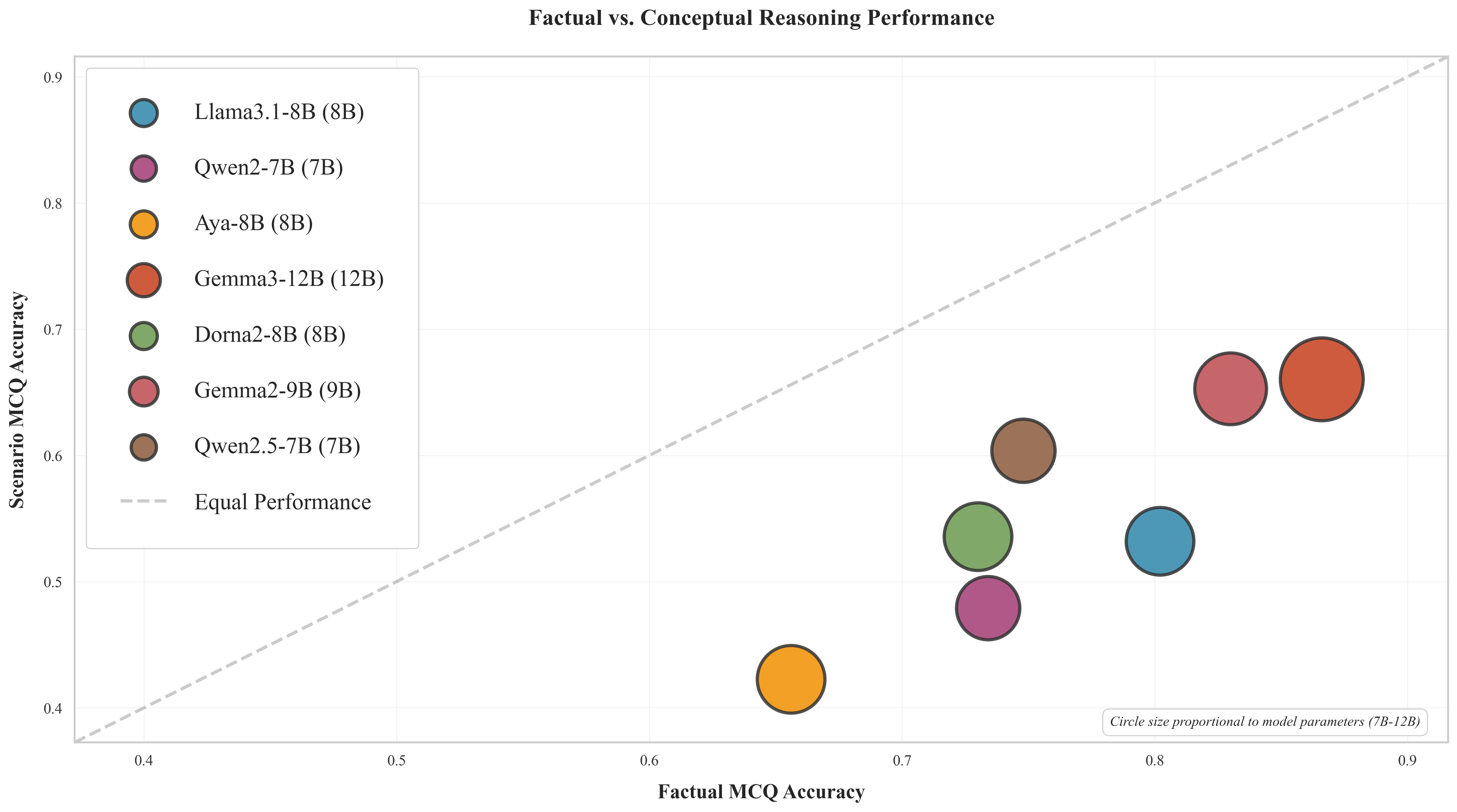}
\caption{Factual Knowledge vs. Cultural Reasoning Gap. Gemma3-12B (largest bubble, top-right) achieves highest factual accuracy but shows inconsistent scenario reasoning. All models fall below the diagonal, indicating systematic difficulty in transferring factual knowledge to cultural schema application. Bubble size represents model parameters.}
\label{fig:factual_vs_scenario}
\end{figure}

\paragraph{Finding 4: Size Does Not Guarantee Cultural Intelligence}

Gemma3-12B, the only model exceeding 10B parameters, shows contradictory performance patterns: it achieves best factual retrieval (87\%, 21\% above average) yet worst acquiescence bias (+61\%, tied with Dorna2). The larger model excels at memorizing cultural facts while showing extreme acquiescence bias, accepting 92\% of positive statements but rejecting only 31\% of negative ones. This suggests that model size improves fact memorization without improving cultural discernment, and larger models may amplify pattern-matching behaviors learned from training data.

%% file: Tables/main_table.tex
\begin{table*}[t]
\centering
\small
\begin{tabular}{l c c c c}
\toprule
\textbf{Model} & \textbf{Factual MCQ} & \textbf{Scenario MCQ} & \textbf{Gap} & \textbf{Acquiescence bias} \\
 & (Knowledge) & (Reasoning) & (Fact - Scenario) & (Accept - Reject) \\
\midrule
Gemma3-12B & \textbf{87.0} $\pm$ 1.0 & \textbf{66.0} $\pm$ 1.0 & 21.0 & {+61.0} \\
Gemma2-9B & 83.0 $\pm$ 0.0 & 65.0 $\pm$ 3.0 & 18.0 & {+41.0} \\
Llama3.1-8B & 80.0 $\pm$ 2.0 & 53.0 $\pm$ 5.0 & 27.0 & \textbf{-10.0} \\
Qwen2.5-7B & 75.0 $\pm$ 1.0 & 60.0 $\pm$ 2.0 & \textbf{15.0} & {+26.0} \\
Qwen2-7B & 73.0 $\pm$ 3.0 & 48.0 $\pm$ 5.0 & 25.0 & {+65.0} \\
Dorna2-8B & 73.0 $\pm$ 2.0 & 54.0 $\pm$ 3.0 & 19.0 & {+61.0} \\
Aya-8B & 66.0 $\pm$ 2.0 & 42.0 $\pm$ 2.0 & 24.0 & {+19.0} \\
\midrule
\textbf{Average} & \textbf{76.7} & \textbf{55.4} & \textbf{21.3} & \textbf{+43.3} \\
\bottomrule
\end{tabular}
\caption{Core performance metrics across all models. \textbf{Gap} measures difficulty transferring factual knowledge to scenario reasoning. \textbf{Acquiescence bias} (Accept - Reject) reveals pattern-matching: positive values (red) indicate models accept appropriate behavior readily but fail to reject violations. Only Llama3.1-8B shows skeptical bias (blue). Percentages shown; standard deviations computed across 5 prompt variations. Models sorted by Factual MCQ performance.}
\label{tab:main_results}
\end{table*}

%% file: Tables/Llama_vs_Dorna.tex
\begin{table}[t]
\centering
\small
\begin{tabular}{l c c c}
\hline
\textbf{Task} & \textbf{Llama3.1-8B} & \textbf{Dorna2-8B} & \textbf{Change} \\
\hline
Reject False & 0.73 & 0.30 & {-43\%} \\
Accept True & 0.63 & 0.91 & {+28\%} \\
Factual MCQ & 0.80 & 0.73 & -7\% \\
Scenario MCQ & 0.53 & 0.54 & +1\% \\
\hline
\end{tabular}
\caption{Direct comparison of Llama3.1-8B (base) and Dorna2-8B (Persian-tuned). Persian pretraining dramatically increased acquiescence while destroying critical reasoning.}
\label{tab:llama_vs_dorna}
\end{table}

%% file: Sections/discussion.tex
\label{sec:discussion}

\paragraph{Why Persian Cultural Concepts Are Hard}

Persian cultural concepts exhibit three critical characteristics that resist pattern-matching. First, they encode implicit social contracts never stated in text: \textit{taarof} requires tracking iteration counts, modeling power asymmetries, and distinguishing ritual from sincerity through multi-turn interaction. Second, they involve context-dependent inversion where identical actions have opposite meanings: paying for a meal is appropriate among peers after negotiation but offensive when done secretly with elders. Third, they conflate factual and normative knowledge: "Is burning \textit{esfand} effective?" (factual, no) versus "Is burning \textit{esfand} appropriate when moving?" (normative, yes). Models trained on factual text systematically conflate these categories.

\paragraph{The Acquiescence Trap Mechanism}

The severe acquiescence bias across six models stems from distributional patterns in Persian training data. Models learn that cultural keywords like \textit{taarof}, \textit{nazri}, and \textit{esfand} co-occur with positive contexts: respectful news coverage, affirmative social media posts, and celebratory descriptions. When encountering these keywords, models trigger acceptance regardless of scenario details. This explains why they accept both correct \textit{taarof} (refuse twice, accept third time) and violations (accept immediately). Without reasoning about iteration logic, they pattern-match to cultural keywords instead.

\paragraph{Why Persian Pretraining Degraded Reasoning}

Continuous Persian pretraining on Dorna2 collapsed rejection accuracy by 43 percentage points while boosting acceptance by 28 points. This asymmetry reveals surface pattern reinforcement: Persian corpora contain overwhelmingly affirmative mentions of traditions, creating distributional bias toward accepting culturally-themed statements. Meanwhile, Llama3.1's skeptical stance (73 percent rejection) likely reflects instruction tuning for critical evaluation. Persian pretraining overrode this skepticism, producing cultural fluency without reasoning. This challenges the assumption that monolingual pretraining improves understanding, it may amplify fluency while degrading discernment.

\paragraph{Factual-Conceptual Gap as Schema Evidence}

The consistent 21 percent performance drop from factual to scenario tasks operationalizes the distinction between cultural facts (retrievable knowledge like "Nowruz marks New Year") and cultural schemas (situational rules like "Would serving \textit{halva} be appropriate here?"). Facts require memorization; schemas require conditional reasoning about context, relationships, and implicit norms. Current training successfully instills facts but fails at schemas, suggesting that adding more factual data cannot close this gap, different learning mechanisms are needed.

%% file: Sections/conclusion.tex
\label{sec:conclusion}

We introduced \datasetname, a benchmark evaluating cultural reasoning in Persian LLMs through 81 concepts and 315 questions across three task types. Evaluation of seven models reveals four key findings: most models exhibit severe acquiescence bias, accepting appropriate behavior while failing to reject violations; continuous Persian pretraining amplified this bias rather than improving reasoning; a consistent 21 percent gap between factual and scenario performance demonstrates that cultural facts and schemas engage different mechanisms; and model size improves memorization but not reasoning. These findings challenge assumptions that scaling monolingual data improves cultural competence and suggest that low-resource languages require more advanced methods, rather than simply training the model on more tokens of the target language. We release our benchmark publicly to enable systematic measurement of cultural understanding, providing a template for evaluation in other languages. Persian culture's synthesis of Zoroastrian, Islamic, and folk traditions makes it a demanding testbed for genuine schema reasoning versus surface pattern-matching. Our work establishes a foundation for building language models that truly comprehend the cultures they serve.

%% file: Sections/limitation.tex
\section*{Limitations}

\paragraph{Model size and scaling regime.}
Our study evaluates a narrow band of \emph{small-to-mid sized} open models (7--12B parameters), selected to enable controlled comparisons across systems with broadly similar capacity and inference cost. This design improves comparability, but it limits what we can conclude about how cultural conceptual reasoning behaves at larger scales (e.g., 30B--70B+), where instruction tuning, longer context windows, or different training mixtures may change both \emph{bias profiles} and \emph{reasoning strategies}. In particular, our observation that models can improve factual recall while still exhibiting accept-over-reject asymmetry may not extrapolate monotonically to substantially larger model families; scaling could either attenuate or amplify the ``acquiescence trap,'' depending on how the model is trained and aligned. Additionally, by keeping model size relatively constant, we cannot disentangle whether certain failures are fundamentally capacity-limited versus primarily data- and objective-driven.

\paragraph{Hand-curated dataset and author priors.}
DivanBench is authored manually based on lived cultural knowledge, which increases scenario realism and insider validity but introduces several curation constraints. First, manual writing inevitably reflects the authors' judgments about what counts as the ``canonical'' interpretation of a concept, which may underrepresent regional, socioeconomic, or generational variation. Second, even with iterative review, scenario phrasing can unintentionally leak cues (e.g., overly salient keywords, unnatural dialogue) that models may exploit via superficial heuristics; especially in paired positive/negative formats. Third, manual coverage is bounded: although we span many (more than 80) concepts, the space of culturally meaningful edge cases is far larger, and some concepts may be overrepresented by the scenarios that are easiest to write unambiguously. Finally, because expert annotation is expensive and time-consuming, we do not incorporate large-scale crowdsourced validation or inter-annotator agreement; thus, while items are intended to be unambiguous to cultural insiders, residual ambiguity cannot be ruled out.

%% file: Appendix/dataset_details.tex
\section{Dataset Details}
\label{sec:appendixData}

\subsection{Detailed Cultural Concept Lists}
\label{app:concepts}

Table~\ref{tab:concepts_detailed} provides comprehensive concept coverage across all seven cultural domains.

\begin{table*}[t]
\centering
\small
\begin{tabular}{lp{12cm}}
\toprule
\textbf{Category} & \textbf{Example Concepts} \\
\midrule
\textbf{Social Etiquette} & \textit{Taarof} (ritual politeness), \textit{Jang-e Hesab} (payment battles), Doorway Deference (older enters first), Three-Times Rule (refuse twice), \textit{Pishkesh} (gift-giving protocol), \textit{Shirini} (sweet bringing obligation), Guest-host dynamics, Shoe removal customs, Greeting hierarchies \\
\midrule
\textbf{Superstitions \& Omens} & \textit{Fal-e Hafez} (Hafez divination), \textit{Fal-Gush} (eavesdropping for omens), Dream interpretation, Itchy palms (money coming), Ear ringing (someone talking about you), Twitching eye (bad omen), Sneezing omens, Bird flight patterns, Broken mirrors, Spilled salt \\
\midrule
\textbf{Apotropaic Rituals} & \textit{Esfand} burning (wild rue fumigation), \textit{Nazar} amulet (evil eye protection), Salt circle protection, \textit{Bismillah} invocations, Garlic hanging, Iron deterrence, Seven knots ritual, Mirror placement, Knife under pillow, Holy verses \\
\midrule
\textbf{Supernatural Beings} & \textit{Jinn} (fire spirits), \textit{Div} (demons), \textit{Pari} (fairies), \textit{Bakhtak} (sleep paralysis demon), \textit{Hamzad} (personal spirit double), \textit{Al} (child-stealing demon), \textit{Ghoul}, Evil eye personification, Ancestor spirits \\
\midrule
\textbf{Nowruz Traditions} & \textit{Haft Sin} (seven S's table), \textit{Chaharshanbe Suri} (fire-jumping), \textit{Haji Firuz} (blackface herald), \textit{Samanu Pazan} (wheat pudding stirring), \textit{Khaneh Tekani} (spring cleaning), \textit{Eidi} (new year gift), \textit{Sizdah Bedar} (13th day picnic), \textit{Sabzeh} growing, \textit{Goldfish} symbolism, New clothes tradition, Elder visitation order, Fire-jumping prayers, \textit{Senjed} (jujube significance), Mirror watching, Egg decoration \\
\midrule
\textbf{Wedding Ceremonies} & \textit{Sofreh Aghd} (marriage spread), Honey ritual, Knife dance, \textit{Kuzeh Shekani} (pot breaking), Sugar cone rubbing, Mirror \& candelabra, Needle-sewing ritual, \textit{Aghd} contract, \textit{Shirini Khoran} (dessert ceremony), Witness requirements, Gold coin showering, Henna night \\
\midrule
\textbf{Taboos} & Whistling at night (attracts \textit{jinn}), Sweeping at night (sweeps away prosperity), Stepping on bread (brings poverty), Cutting nails at night (invites demons), Giving knives as gifts (cuts relationships), Opening umbrellas indoors, Haircut on Wednesdays, Shoe upside-down, Passing over children (stunts growth), Trimming nails over water, Direct compliments (causes evil eye), Singing while eating \\
\bottomrule
\end{tabular}
\caption{Comprehensive cultural concept coverage. Each concept has paired positive/negative binary scenarios plus representation in factual or scenario MCQs.}
\label{tab:concepts_detailed}
\end{table*}

\subsection{Additional Examples}
\label{app/samples}

\begin{tcolorbox}[colback=blue!5, colframe=blue!40!black, title=Example: Factual MCQ, fonttitle=\bfseries, top=2mm, bottom=2mm, left=3mm, right=3mm]

\textit{What fruit is used as a symbol on the Yalda night table?}\\
(A) Orange \quad (B) Apple \quad (C) Pomegranate \quad (D) Pear

\textbf{Expected}: (C) Pomegranate

\end{tcolorbox}

\begin{tcolorbox}[colback=blue!5, colframe=blue!40!black, title=Example: Binary Belief Verification, fonttitle=\bfseries, top=2mm, bottom=2mm, left=3mm, right=3mm]

\textbf{Positive}: ``During a casual family gathering, an elderly aunt walks into the room. Everyone younger immediately stands up as a respectful acknowledgment of her presence. Did they act according to Persian tradition?''

\textbf{Expected}: Yes (This follows the rule to stand up when elders enter)

\vspace{2mm}

\textbf{Negative}: ``During a family gathering, an elderly aunt enters the room. Since everyone is already comfortable, no one stands up and they simply greet her warmly from their seats. Did they act according to Persian tradition?''

\textbf{Expected}: No (This violates the rule to stand up when elders enter)

\end{tcolorbox}

%% file: Appendix/experiment_details.tex
\section{Appendix}
\label{sec:extra-exp}
\subsection{Model Descriptions}
\label{app:models}

\input{Tables/detailed_table_appendix}

\paragraph{Aya-8B}
Multilingual model from CohereForAI with Persian support. Trained on diverse multilingual data including Persian web text and translated datasets. Instruction-tuned for helpfulness.

\paragraph{Dorna2-8B}

Persian-optimized variant of Llama3.1-8B created through continuous pretraining on Persian corpora including news, social media, literature, and web text. Maintained Llama3.1's instruction-tuning while adapting vocabulary and linguistic patterns to Persian.

\paragraph{Gemma2-9B and Gemma3-12B}

Google's open-source multilingual foundation models (second and third generations). Trained on diverse web data including Persian content. Gemma3 represents scaled-up version with 12B parameters.

\paragraph{Llama3.1-8B}
Meta's base model serving as foundation for Dorna2. Multilingual capabilities including Persian from pretraining on web-scale data. Instruction-tuned for general helpfulness and reasoning.

\paragraph{Qwen2-7B and Qwen2.5-7B}
Alibaba's multilingual models (second generation and improved variant). Strong performance on Asian languages including Persian. Qwen2.5 incorporates improved training procedures and data quality.

\subsection{System Prompt Variations}
\label{app:prompts}

We use 5 diverse system prompts to ensure robust evaluation and they are shared among all tasks. Examples for binary verification tasks:

\begin{tcolorbox}[colback=gray!5, colframe=gray!40!black, title=System Prompt 1, fonttitle=\bfseries, top=2mm, bottom=2mm, left=0mm, right=0mm]

``You are a helpful assistant who is an expert in Iranian culture, folklore, traditions, and customs.''

\end{tcolorbox}

\begin{tcolorbox}[colback=gray!5, colframe=gray!40!black, title=System Prompt 2, fonttitle=\bfseries, top=2mm, bottom=2mm, left=0mm, right=0mm]

``You are a knowledgeable cultural expert specializing in Persian traditions, superstitions, and folk beliefs. Answer questions based on traditional Iranian cultural knowledge.''

\end{tcolorbox}

\begin{tcolorbox}[colback=gray!5, colframe=gray!40!black, title=System Prompt 3, fonttitle=\bfseries, top=2mm, bottom=2mm, left=0mm, right=0mm]

``You are an assistant with deep knowledge of Iranian heritage, including traditional customs, superstitions, proverbs, and cultural practices passed down through generations.''

\end{tcolorbox}

\begin{tcolorbox}[colback=gray!5, colframe=gray!40!black, title=System Prompt 4, fonttitle=\bfseries, top=2mm, bottom=2mm, left=0mm, right=0mm]

``As an expert in Iranian folklore and cultural traditions, provide accurate answers about Persian customs, beliefs, superstitions, and traditional practices.''

\end{tcolorbox}

\begin{tcolorbox}[colback=gray!5, colframe=gray!40!black, title=System Prompt 5, fonttitle=\bfseries, top=2mm, bottom=2mm, left=0mm, right=0mm]

``You are a cultural historian specializing in Iranian and Persian traditions. Draw upon your extensive knowledge of Iranian superstitions, customs, ceremonies, and folk beliefs to answer questions.''

\end{tcolorbox}

\subsection{Binary Classification Task complementary Results}
Exact numbers for figure \ref{fig:agreeing_bias}
 are given in table \ref{tab:binary_performance}.

%% file: Tables/detailed_table_appendix.tex
\begin{table*}[t]
\centering
\small
\begin{tabular}{l c c c}
\toprule
\textbf{Model} & \textbf{Accept True (Positive)} & \textbf{Reject False (Negative)} & \textbf{Bias} \\
\midrule
Qwen2-7B & 84.0 $\pm$ 4.0 & 19.0 $\pm$ 1.0 & +65.0 \\
Gemma3-12B & 92.0 $\pm$ 3.0 & 31.0 $\pm$ 3.0 & +61.0 \\
Dorna2-8B & 91.0 $\pm$ 5.0 & 30.0 $\pm$ 5.0 & +61.0 \\
Gemma2-9B & 84.0 $\pm$ 2.0 & 43.0 $\pm$ 7.0 & +41.0 \\
Qwen2.5-7B & 74.0 $\pm$ 7.0 & 48.0 $\pm$ 6.0 & +26.0 \\
Aya-8B & 57.0 $\pm$ 4.0 & 38.0 $\pm$ 5.0 & +19.0 \\
Llama3.1-8B & 63.0 $\pm$ 7.0 & 73.0 $\pm$ 5.0 & -10.0 \\
\bottomrule
\end{tabular}
\caption{Binary verification task performance sorted by bias magnitude. Models show severe asymmetry: high acceptance of appropriate behavior but low rejection of violations, indicating surface-level pattern-matching.}
\label{tab:binary_performance}
\end{table*}